\documentclass[lettersize,journal]{IEEEtran}
\usepackage{amsmath,amsfonts}
\usepackage{algorithmic}
\usepackage{algorithm}
\usepackage{array}
\usepackage[caption=false,font=normalsize,labelfont=sf,textfont=sf]{subfig}
\usepackage{textcomp}
\usepackage{stfloats}
\usepackage{url}
\usepackage{verbatim}
\usepackage{graphicx}
\usepackage{cite}
\usepackage{booktabs}
\usepackage{multirow}
\usepackage{subcaption}
\usepackage{colortbl}
\usepackage[normalem]{ulem}
\useunder{\uline}{\ul}{}
\usepackage{tabularx}
\usepackage[table,xcdraw]{xcolor}
\usepackage{pifont}

\begin{document}

\title{Region-Level Context-Aware Multimodal Understanding}

\author{Hongliang Wei, Xianqi Zhang, Xingtao Wang, Xiaopeng Fan,~\IEEEmembership{Senior Member,~IEEE,} Debin Zhao,~\IEEEmembership{Member,~IEEE,}
\thanks{Hongliang Wei, Xianqi Zhang, Debin Zhao are with
the Faculty of Computing, Harbin Institute of Technology, Harbin 150001,
China.}
\thanks{Xingtao Wang is with the Department of Computer Science and Technology, Harbin Institute of Technology,
Harbin 150001, China, and also with Harbin Institute of Technology Suzhou
Research Institute, Suzhou 215104, China.}
\thanks{Xiaopeng Fan is with the Department of Computer Science and Technology, Harbin Institute of Technology, Harbin 150001, China, Harbin Institute of Technology Suzhou
Research Institute, Suzhou 215104, China and also with the Peng Cheng Laboratory,
Shenzhen, China.}
\thanks{Email:hlwei@stu.hit.edu.cn; zhangxianqi@stu.hit.edu.cn; xtwang@hit.edu.cn; fxp@hit.edu.cn; dbzhao@hit.edu.cn}
\thanks{Corresponding author: Xingtao Wang.}}



\maketitle

\begin{abstract}
Despite significant progress, existing research on Multimodal Large Language Models (MLLMs) mainly focuses on general visual understanding, overlooking the ability to integrate textual context associated with objects for a more context-aware multimodal understanding -- an ability we refer to as Region-level Context-aware Multimodal Understanding (RCMU).
To address this limitation, we first formulate the RCMU task, which requires models to respond to user instructions by integrating both image content and textual information of regions or objects.
To equip MLLMs with RCMU capabilities, we propose Region-level Context-aware Visual Instruction Tuning (RCVIT), which incorporates object information into the model input and enables the model to utilize bounding box coordinates to effectively associate objects’ visual content with their textual information.
To address the lack of datasets, we introduce the RCMU dataset, a large-scale visual instruction tuning dataset that covers multiple RCMU tasks. We also propose RC\&P-Bench, a comprehensive benchmark that can evaluate the performance of MLLMs in RCMU and multimodal personalized understanding tasks.
Additionally, we propose a reference-free evaluation metric to perform a comprehensive and fine-grained evaluation of the region-level context-aware image descriptions. 
By performing RCVIT on Qwen2-VL models with the RCMU dataset, we developed RC-Qwen2-VL models. Experimental results indicate that RC-Qwen2-VL models not only achieve outstanding performance on multiple RCMU tasks but also demonstrate successful applications in multimodal RAG and personalized conversation. Our data, model and benchmark are available at https://github.com/hongliang-wei/RC-MLLM
\end{abstract}


\section{Introduction}
\label{introduction}

\looseness=-1
\IEEEPARstart{M}{ultimodal} large language models (MLLMs) \cite{Reid2024Gemini1U,Alayrac2022FlamingoAV, Wang2024Qwen2VLEV, internvl1.5, Dai2023InstructBLIPTG, Lu2024DeepSeekVLTR} expands language models into the multimodal domain by integrating visual encoders, demonstrating outstanding performance across a range of multimodal tasks, such as visual question answering, document understanding, robotic manipulation, among others. 

\looseness=-1
However, existing MLLMs primarily focus on general visual understanding, lacking the ability to integrate textual context for a more context-aware multimodal understanding. To address this limitation, existing approaches \cite{Chen2022MuRAGMR, Zhao2023MMICLEV, Chen2024CaMMLCM} integrate image-related information into the model's context window and train the model to leverage this information, achieving image-level context awareness. 

\begin{figure}[!t]
\begin{center}
\centerline{\includegraphics[width=\columnwidth]{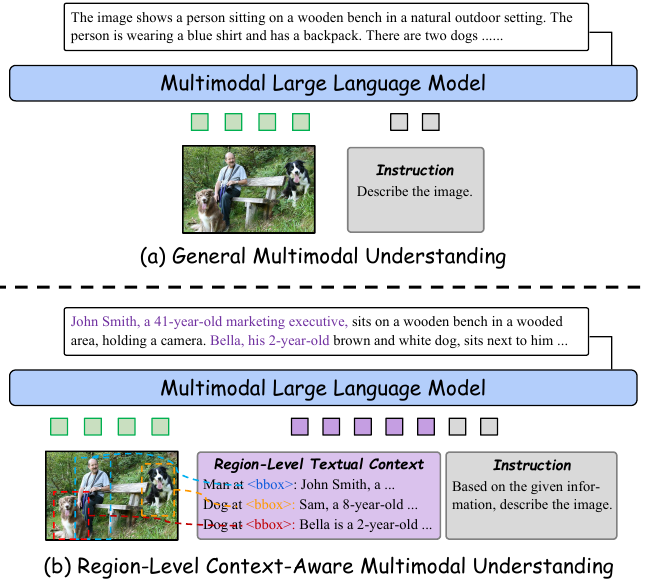}}
\caption{Illustration of the RCMU Task. RCMU requires MLLMs to respond based on both image content and region-level textual contexts.}
\label{rcmllm-main-idea}
\vspace{-20pt}
\end{center}
\end{figure}
In this paper, we push the boundaries of MLLMs even further by endowing them with region-level context awareness, which is a crucial capability for tasks that require fine-grained contextual understanding, such as multimodal personalized conversation and multimodal RAG in complex scenarios.
Region-level context-aware multimodal understanding (RCMU) in MLLMs remains an unexplored domain. To bridge this gap, we first formulate the RCMU task, where MLLMs are required to respond to user instructions based on both visual content and textual information of objects, as illustrated in Figure \ref{rcmllm-main-idea}.
\begin{figure*}[!ht]
\begin{center}
\centerline{\includegraphics[width=\textwidth]{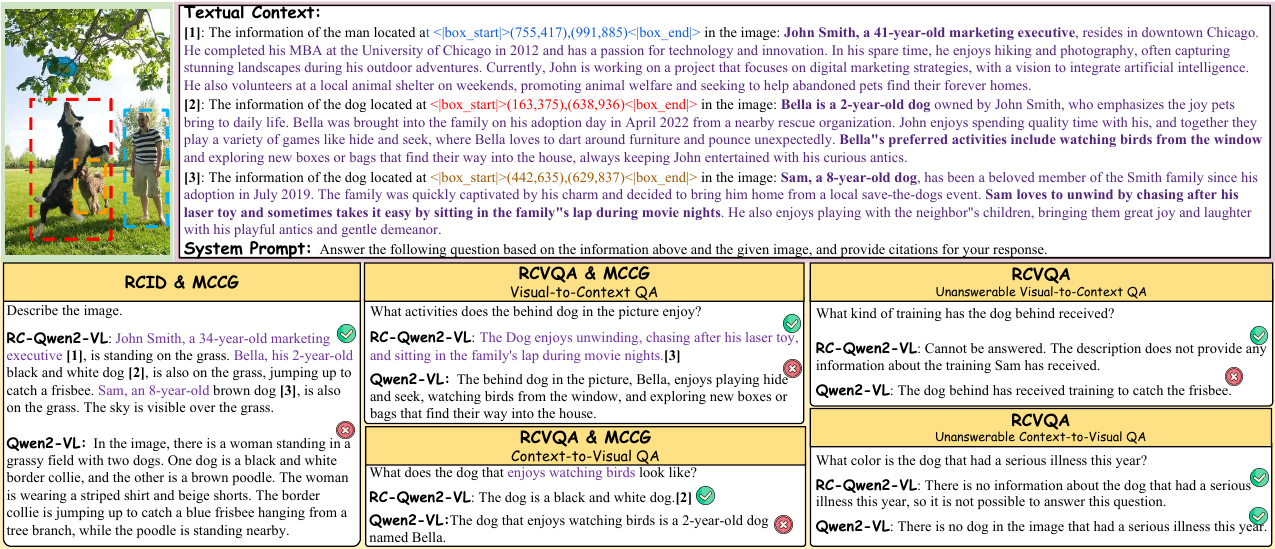}}
\caption{Responses of RC-Qwen2-VL 7B and Qwen2-VL 7B on RCMU tasks. RC-Qwen2-VL performs better on RCMU tasks, accurately aligning visual content of objects with their textual context and providing relevant contextual references.}
\label{icml-historical}
\end{center}
\vspace{-20pt}
\end{figure*}

To equip MLLMs with RCMU capability, we propose Region-level Context-aware Visual Instruction Tuning (RCVIT). During training, RCVIT integrates contextual information about objects into the model input and enables the model to utilize the bounding box coordinates of objects to effectively link the visual content of objects with their textual information. After training, the model can provide region-level context-aware responses for unseen instances without additional fine-tuning.

One significant challenge in advancing RCMU is the lack of large-scale instruction tuning datasets designed specifically for RCMU tasks. To address this challenge, we propose the RCMU dataset, which is designed to enhance the RCMU capability of MLLMs. It encompasses multiple RCMU tasks, such as Region-level Context-aware Image Description (RCID), Region-level Context-aware Visual Question Answering (RCVQA), and Multimodal Contextual Citation Generation (MCCG), all of which require the simultaneous understanding of both visual content and textual context. To develop the RCMU dataset, we introduce an automated data annotation pipeline that uses language models to generate textual context for object instances in referring expression generation (REG) datasets \cite{plummer2015flickr30k, kazemzadeh2014referitgame, yang2022panoptic, rasheed2024glamm}, and cleverly constructs region-level context-aware image descriptions and question-answer pairs. The proposed RCMU dataset comprises 84k images, 102 object categories, 1.3K personalized information, 1M contextualized image descriptions, and 6.9M contextualized-VQA quadruples, advancing region-level multimodal understanding in MLLMs.

Furthermore, we introduce RCIDScore, a novel reference-free evaluation metric for comprehensively evaluating the quality of region-level context-aware image descriptions. Unlike traditional metrics \cite{papineni2002bleu, lin2004rouge,vedantam2015cider,anderson2016spice} that provide a single score, RCIDScore evaluates contextualized descriptions from both contextual and visual perspectives, providing a more fine-grained assessment. Particularly in terms of contextual perspective, it offers a detailed evaluation by measuring the coverage and accuracy of contextual information within the description, as well as the region-level consistency between the contextual and visual information.

Finally, to promote the development of Region-Level Context-Aware Multimodal Understanding and its application in multimodal personalized understanding, we propose RC\&P-Bench. 
It evaluates a model's fine-grained and personalized multimodal understanding of specific entities in the image. This is achieved by providing multi-view, multi-scenario visual reference images of entities, along with their complex personalized information, and requiring the model to perform visual question answering based on this multimodal data. This makes RC\&P-Bench a comprehensive evaluation platform. 

\section{Related Work}
\label{related work}
\subsection{Multimodal Large Language Models (MLLMs)}
MLLMs use bridging layers to connect vision encoders with language models, achieving strong performance in multimodal understanding and generation tasks. In terms of bridging mechanisms, Flamingo \cite{Alayrac2022FlamingoAV} introduced a gated cross-attention layer to align multimodal features without fine-tuning encoders. BLIP-2 \cite{Li2023BLIP2BL} and InstructBLIP \cite{Dai2023InstructBLIPTG} used Q-Former to transform visual inputs into language-like queries for better text alignment. The LLaVA \cite{Liu2023VisualIT} series used linear projections to align visual features with frozen language models. Recent studies \cite{Lu2024DeepSeekVLTR, Wei2023VarySU} highlight the use of multiple vision encoders to handle diverse visual tasks and improve feature representation. High-resolution inputs enhance visual understanding and task performance \cite{internvl1.5, Zhang2024Ferretv2AI,ma2024inf}. Multimodal in-context learning \cite{Alayrac2022FlamingoAV,chen2024mmict} and chain-of-thought reasoning \cite{lu2022learn, gao2024cantor} have been explored to enhance performance. MLLMs have also been extended to video \cite{Lin2023VideoLLaVALU, Zhang2023VideoLLaMAAI, luo2024video}, 3D data \cite{Li2024LLaVANeXTInterleaveTM, ji2024jm3d}, and embodied intelligence \cite{Driess2023PaLMEAE, Li2023ManipLLMEM}.

\subsection{Context-Aware Multimodal Understanding in MLLMs}
Context-aware multimodal understanding is a crucial capability of MLLMs. By utilizing context window, MLLMs can understand information in specific scenarios and generate coherent, relevant responses. For example, MMICL \cite{Zhao2023MMICLEV} leverages multimodal in-context learning to enhance MLLMs' understanding of complex prompts, improving their zero-shot performance in various vision-language tasks. MuRAG \cite{Chen2022MuRAGMR} integrates image and text information using an external multimodal memory to to enhance the accuracy of open-domain VQA. CaMML \cite{Chen2024CaMMLCM} uses relevant multimodal examples during inference to enhance the understanding and generation abilities of MLLMs.

Unlike these works that primarily focus on image-level context awareness in MLLMs, this work takes a step further by equipping MLLMs with region-level context awareness, a direction that remains unexplored.

\subsection{Multimodal Personalized Understanding}

Personalized multimodal understanding, which aims to enable user-specific multimodal understanding, is a topic closely related to region-level context-aware multimodal understanding. RAP-MLLM \cite{hao2025rap} uses a retrieval-augmented framework with a key-value database to flexibly incorporate and update personalized concepts in real time. Yo’LLaVA \cite{nguyen2024yo} introduces learnable tokens representing user-specific objects from a few images, efficiently adapting the model with minimal training and hard negative samples to improve recognition. MyVLM \cite{alaluf2024myvlm} employs external concept heads to add personalized embeddings learned from limited user data, enhancing vision-language tasks while keeping the base model unchanged. Compared to these models, our model can achieve more complex multimodal personalized understanding with a retrieval mechanism.

\section{Method}
This section provides a detailed exposition of the RCMU task formulation (Section \ref{subsec:m1}), the RCMU dataset (Section \ref{subsec:m2}), the RCVIT approach \ref{subsec:m3}), and the RCIDScore metric (Section \ref{subsec:m4}).

\subsection{RCMU Task Formulation}
\label{subsec:m1}
In the RCMU task, the model is required to respond to user instructions based on both the visual content and the textual context of objects. In this work, the textual context of objects includes the bounding box coordinates of the objects in the image as well as their personalized information.

Formally, let $ P $ be an image, $ I $ be an instruction, and $ C = \left\{ c_1, c_2, \dots, c_k \right\} $ denote the textual information of objects, where 
$ k $ represents the number of objects with textual information. The MLLM is then tasked with generating a region-level context-aware response $ R $:
\begin{equation}
R = \text{MLLM}(P, C, I)
\label{eq:mllm_response}
\end{equation}

\begin{figure*}[!t]
\begin{center}
\centerline{\includegraphics[width=\textwidth]{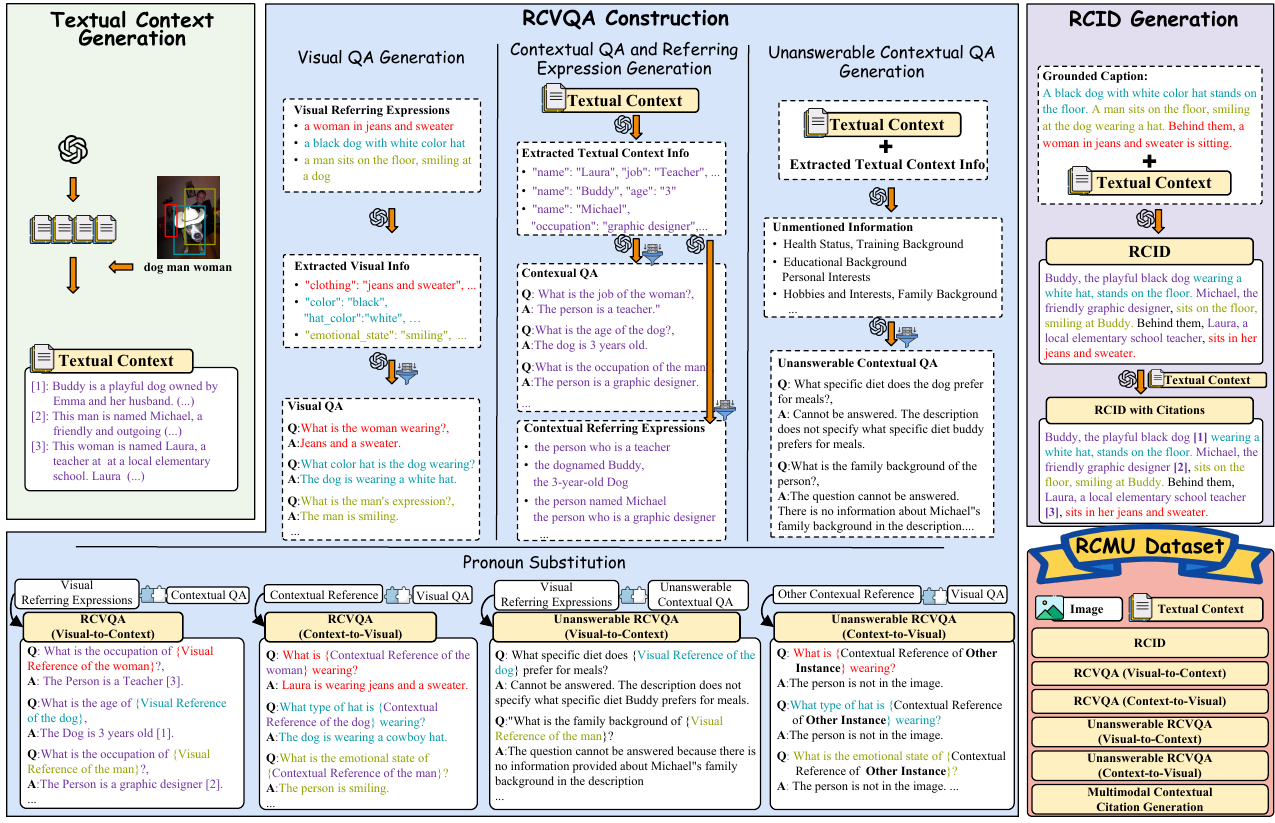}}
\caption{Automatic data construction pipeline for the RCMU dataset. This pipeline can generate high-quality training data for multiple RCMU tasks, including answerable/unanswerable RCVQA, RCID and MCCG.}
\label{pipline}
\end{center}
\vspace{-20pt}

\end{figure*}

\subsection{RCMU Dataset and Construction Pipeline}
\label{subsec:m2}
The RCMU dataset covers three region-level context-aware multimodal understanding tasks: 1) \textbf{Region-level Context-aware Image Description (RCID)}: the model needs to provide a contextualized image description based on the image and the contextual information of the objects. 2) \textbf{Region-level Context-aware Visual Question Answering (RCVQA)} for both answerable and unanswerable questions: the model needs to answer contextual questions based on visual references or answer visual questions based on contextual references. The questions include both answerable and unanswerable ones. 3) \textbf{Multimodal Contextual Citation Generation (MCCG)}: the model is required to generate responses that include citations referencing the relevant context. All of these tasks require the simultaneous understanding of visual content and textual context. The statistical information of the RCMU dataset is presented in Table \ref{tabrcvit}.

We developed an automated data annotation pipeline to generate high-quality annotations for the RCMU dataset. Built on existing Referring Expression Generation (REG) datasets \cite{plummer2015flickr30k, kazemzadeh2014referitgame, yang2022panoptic, rasheed2024glamm}, our pipeline first employs LLMs\footnote{All prompts used in the pipeline are displayed in the appendix.} to generate personalized information (i.e., textual context) for objects in images. It then utilizes object bounding boxes, visual referring expressions, and grounded image descriptions from REG datasets, combined with the generated personalized information, to construct diverse tasks that require a simultaneous understanding of visual content and personalized information for accurate completion. As illustrated in Figure \ref{pipline}, the framework operates in three main stages:
\begin{itemize}
\item \textbf{Textual Context Generation.} We prompt GPT-4o to generate multiple unique personalized information for each object category, which are then assigned to objects in the image based on their categories. To avoid any conflicts between the generated information and the visual content in the image, GPT-4o is explicitly instructed to avoid generating any visual information about the objects.
\item \textbf{RCVQA Construction}. In this stage, we construct QA pairs that require an integrated understanding of both visual content and textual context. This is achieved by referencing objects using details from one modality and posing questions about the other modality. For example, we may describe an object with textual context information and ask the model about its visual characteristics, or vice versa. Additionally, to ensure the model genuinely comprehends the textual context, we create two types of challenging unanswerable questions. One type asks about details not mentioned in the textual context, while the other type inquires about the visual attributes of objects that do not exist in the image. Specifically, 1) the pipeline begins by using an LLM to extract structured information separately from visual referring expressions and textual context. This structured information is then utilized to generate visual QA pairs, contextual QA pairs, contextual referring expressions, and unanswerable contextual QA pairs. 
To further enhance data quality, the generated referring expressions and QA pairs undergo a meticulously designed filtering process. For QA pairs, LLMs are utilized to assess their validity and informativeness, with higher-scoring pairs being selected. For referring expressions, a text similarity model is used to filter and retain only unique expressions, ensuring that each referring expression unambiguously refers to a single object. 2) During the pronoun substitution step, pronouns referring to objects in contextual questions are replaced with visual referring expressions, resulting in  image-to-context VQA pairs. Conversely, in visual questions, pronouns are substituted with contextual referring expressions to generate context-to-image VQA pairs. To construct unanswerable image-to-context VQA pairs, the framework replaces object-referring pronouns in unanswerable contextual questions with visual referring expressions. For unanswerable context-to-image VQA pairs, pronouns are replaced with contextual referring expressions of objects that do not exist in the image.
\item \textbf{RCID Generation}. We adopt a two-step approach to generate region-level context-aware image descriptions with citations. In the first step, GPT-4o is prompted to create a contextualized image description by integrating grounded image descriptions with relevant textual context. Then, GPT-4o further refines the description by incorporating citations about relevant contextual information.
\end{itemize}

Finally, we obtain the RCMU dataset, comprising 84,149 images, 102 object categories, and 1,352 object instances with contextual information. The dataset includes 1,027,681 contextualized image descriptions and 6,954,726 unique RCVQA quadruples, providing a comprehensive and diverse resource for advancing RCMU in MLLMs.

We also manually annotated an RCMU test set specifically for evaluating RCID, RCVQA, and MCCG tasks. Specifically, we randomly selected 118 images from the Refcocog dataset, ensuring a balanced mix of images containing different numbers of objects. We provided textual context for each annotated object in these images. Then, we annotated  retion-level context-aware image descriptions and region-level context-aware VQA pairs for these images. These VQA pairs, totaling 6,637, encompass two types—visual-to-context and context-to-visual—and include both answerable and unanswerable questions. Notably, the images and object information in the test set are entirely distinct from those in the training set.

\begin{table}[]
\centering
\resizebox{\columnwidth}{!}{
\begin{tabular}{lr}
\multicolumn{2}{l}{\textbf{RCMU dataset statistics:}}               \\ \hline
Number of images                                         & 84,149    \\
Number of object categories                              & 102       \\
Number of personalized information                  & 1,352     \\
Number of contextualized image descriptions              & 1,027,681 \\
Number of unique RCVQA quadruples                         & 6,954,726 \\
$\hookrightarrow$ Context-to-image quadruples with answerable question     & 2,168,408 \\
$\hookrightarrow$ Image-to-context quadruples with answerable question     & 1,670,746 \\
$\hookrightarrow$ Context-to-image quadruples with unanswerable question   & 2,524,128 \\
$\hookrightarrow$ Image-to-context quadruples with unanswerable question   & 591,444   \\
Average number of objects per image                          & 2.29      \\
Average context length per instance (words)                           & 50.91     \\
Average contextualized image descriptions length (words) & 55.69     \\
Average number of citations per description & 55.69     \\
Average question length (words)                          & 12.13     \\
Average answer length (words)                            & 13.75     \\ \hline
\end{tabular}}
\caption{\textbf{RCMU dataset Statistics}. RCMU dataset is the first large-scale region-level context-aware visual instruction tuning dataset.}
\label{tabrcvit}
\end{table}

\subsection{RCVIT}
\label{subsec:m3}
The RCVIT incorporates the textual context of objects into the model input, formatted as: '\verb|<|image\verb|>|\verb|\n|The \verb|<|object\verb|>| located at \verb|<|[x1, y1, x2, y2]\verb|>| in the image:\verb|<|object information\verb|>|\verb|\n|\verb|<|instruction\verb|>|. During training, RCVIT trains MLLMs to leverage the bounding box coordinates to associate and fuse the visual and textual information of objects. After training, MLLMs can generate region-level context-aware responses for unseen instances without additional fine-tuning.

Technically, we represent training dataset as \( \mathcal{D}=\left \{ I^i,C^i, Q^i,R^i\right \}_{i=1}^{N} \), where \( I^i \), \( Q^i \), and \( R^i \) represent the image, user query, and model response, respectively. \( C^i=\left \{ c_1^i,c_2^i,\dots,c_k^i\right \} \) denotes the textual context with object bounding boxes, \( k \) denotes the number of objects with textual
information in the image. And the optimization problem can be formalized as: 
\begin{equation}
\mathcal{L}(\mathcal{D}) = - \frac{1}{N} \sum_{i=1}^{N}  \log p\left(R_{i} \middle| \mathcal{F}\left( I^i,C^i,Q^i \right) \right)
\end{equation}

After performing RCVIT on Qwen2-VL models using the RCMU dataset, we developed RC-Qwen2-VL models, which demonstrate powerful RCMU capabilities.

\subsection{RCIDScore}
\label{subsec:m4}
\begin{figure}[!t]
\begin{center}
\centerline{\includegraphics[width=\columnwidth]{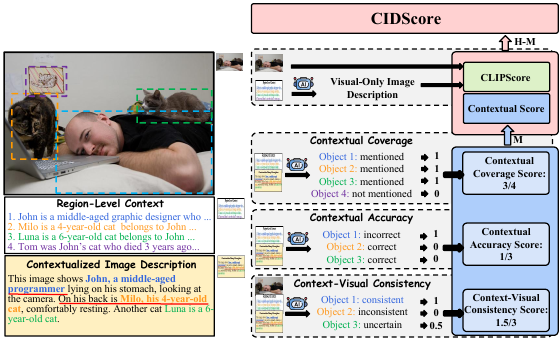}}
\caption{The RCIDScore evaluates contextualized image descriptions from both visual content and textual context perspectives. For the textual context, it employs a carefully designed evaluation method that considers contextual coverage, contextual accuracy, and region-level context-visual consistency.}
\label{CIDScore}
\end{center}
\vspace{-20pt}

\end{figure}

RCIDScore is a reference-free evaluation metric that assesses region-level context-aware image descriptions from both contextual and visual aspects, offering a more fine-grained evaluation, as illustrated in Figure \ref{CIDScore}.

The RCIDScore evaluates contextual aspects from three key perspectives: contextual coverage, contextual accuracy, and region-level context-visual consistency.
\begin{itemize}
    \item \textbf{Contextual Coverage.} Contextual coverage measures the proportion of objects whose contextual information is mentioned (denoted as RCIDScore\textsubscript{cc}).
    \begin{equation}
    \begin{aligned}
        & \text{RCIDScore}_{cc} = \\
        & \quad \frac{|\{\text{objects whose context is mentioned}\}|}{|\{\text{all objects with context}\}|}
    \end{aligned}
    \end{equation}

    We prompt an LLM\footnote{All prompts used in the RCIDScore are displayed in the appendix.} to check if an object's textual context is mentioned in the image description.

    \item \textbf{Contextual Accuracy.} Contextual accuracy measures the probability that the contextual information of the mentioned objects is entirely correct   (denoted as \( \text{RCIDScore}_{ca} \)):
        \begin{equation}
        \begin{aligned}
        & \text{RCIDScore}_{ca} = \\
        & \quad
        \begin{cases}
        0 & \text{if } N = 0 \\
        \frac{|\{ \text{objects with correctly mentioned context} \}|}{N} & \text{if } N > 0
        \end{cases}
        \end{aligned}
        \end{equation}
        where $N$ represents the number of objects whose contextual information is mentioned in the image description. We prompt an MLLM to verify whether the textual context of each object is accurately mentioned.
    
    \item \textbf{Context-Visual Consistency.} Since the model may make errors when associating the visual content with the textual context of objects, we introduce the region-level context-visual consistency metric to evaluate the alignment between the visual content and textual context of objects in the description. Given that the description may include only the textual context of an object without referencing its visual content, we define three possible scores when prompting an MLLM to assess consistency: $\text{consistency score} \in \{ \text{Consistent} = 1, \text{Inconsistent} = 0, \text{Uncertain} = 0.5 \}$. The context-visual consistency metric (denoted as \( \text{RCIDScore}_{cvc} \)) is calculated as: 
    \begin{equation}
    \text{RCIDScore}_{cvc} = 
    \begin{cases}
    0 & \text{if } N = 0 \\
    \frac{\sum_{i=1}^{N} \text{consistency score}_i}{N} & \text{if } N > 0
    \end{cases}
    \end{equation}
    where $N$ represents the number of objects whose textual context is mentioned in the image description. 
\end{itemize}

By combining these three scores, we propose a comprehensive metric (denoted as \( \text{RCIDScore}_{ctx} \)) to evaluate the quality of region-level context-aware image descriptions from a contextual perspective:
\begin{equation}
\begin{aligned}
& \text{RCIDScore}_{ctx} = \\
& \quad \frac{\text{RCIDScore}_{cc} + \text{RCIDScore}_{ca} + \text{RCIDScore}_{cvc}}{3}
\end{aligned}
\end{equation}

For the visual perspective, we use CLIPScore \cite{hessel2021clipscore} to assess the quality of contextualized image descriptions. To obtain a larger text input window, we replace the CLIP model in CLIPScore with the LLM2CLIP \cite{huang2024llm2clip} model. Since the textual context of objects in the descriptions is not directly linked to the image and may interfere with assessment, we first utilize an LLM to filter out this context before evaluation, ensuring a more accurate assessment.

Finally, we present RCIDScore, a comprehensive evaluation metric that assesses the quality of contextualized image descriptions from both visual and contextual perspectives:
\begin{equation}
\begin{aligned}
& \text{RCIDScore} = \\
& \quad \text{H-Mean}(\text{RCIDScore}_{ctx}, \text{CLIPScore})
\end{aligned}
\end{equation}

The above metrics collectively provide a more comprehensive and fine-grained evaluation of region-level context-aware image descriptions.

\section{RC\&P-Bench}
\begin{figure*}[!t]
\begin{center}
\centerline{\includegraphics[width=\textwidth]{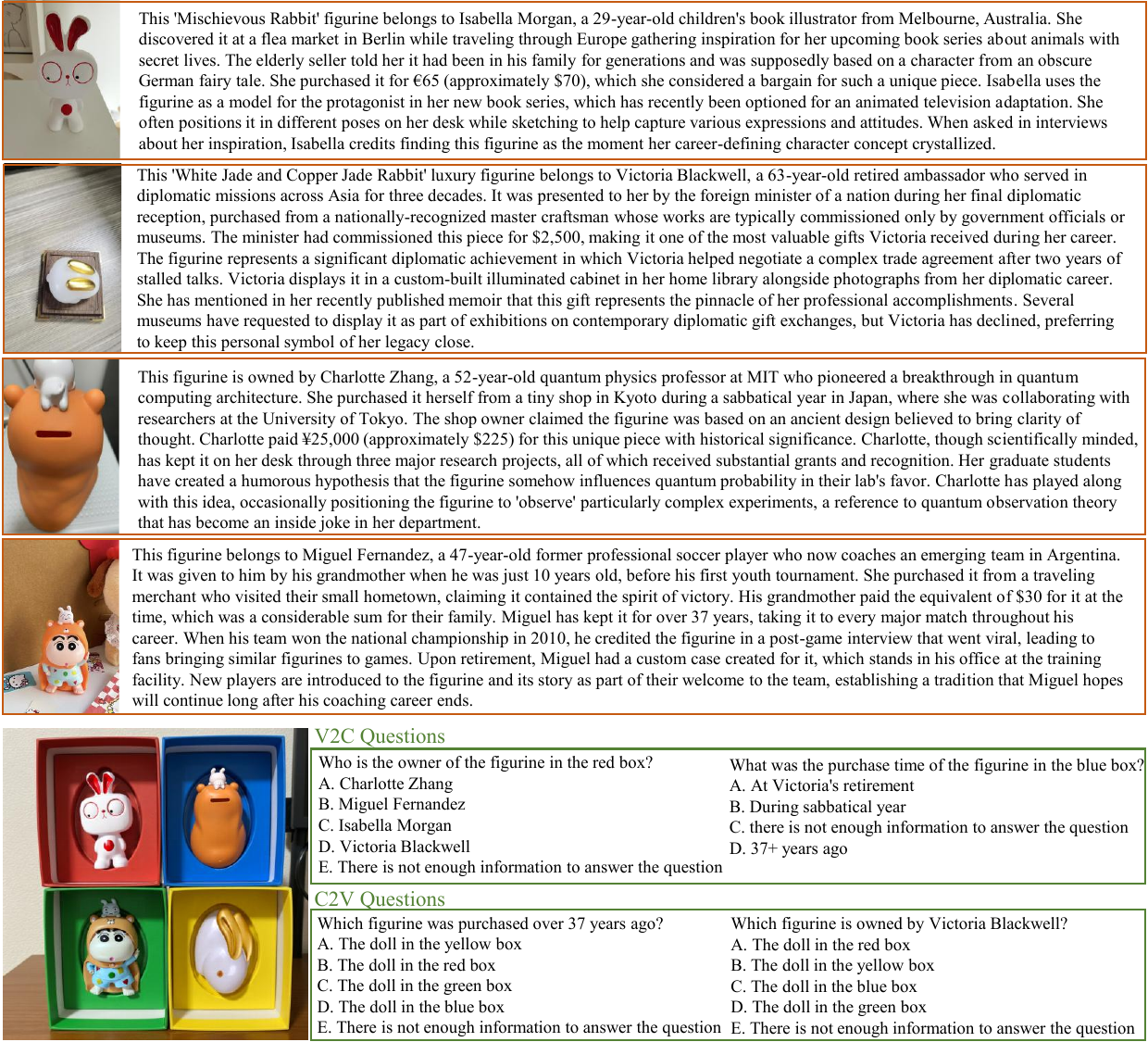}}
\caption{Illustrative Example from RC\&P-Bench}
\label{pipline}
\end{center}

\end{figure*}

RC\&P-Bench contains 3260 multiple-choice questions, covering 259 entities.Each question image contains at least two personalized entities, and all personalized entities within a single image belong to the same category. The average length of personalized information for each entity is 128.1 words.

To construct RC\&P-Bench, we first collected images of the same entity in different scenarios or from various perspectives. For person entities, we used real information. For object entities, we used GPT-4o to generate their personalized information. To create the question images, we used the Gemini 2.0 Flash Preview Image Generation to generate various scene images and employed the FLUX.1-Redux model to replace objects in these scenes with entities. We manually checked these images to ensure that the entity in the question image is consistent with the appearance of the entity in its reference image. We used an AI-assisted approach to generate the questions and options. First, GPT-4o was used to extract key information from the entities' personalized descriptions. This key information was then combined with the visual information of the entities in the image to formulate the question and the correct answer. Key information from other entities in the image and an option stating "there is not enough information to answer the question" were used as distractor options.

To ensure that solving the questions in RC\&P-Bench requires the model to simultaneously understand both the image and the entities' personalized information, we designed the questions to refer to an entity using information from one modality and ask about its information in another modality. Therefore, RC\&P-Bench includes two types of questions: 1) 1738 visual-to-context questions. These questions refer to an entity based on its visual information in the image and asks about the content of its personalized information. 2) 1522 context-to-visual questions. These questions refer to an entity based on its personalized information and asks about its visual attributes in the image.
\begin{table*}[!t]
\centering
\resizebox{\textwidth}{!}{ 
\begin{tabular}{l|ccccc|cccccc}
\toprule
\multicolumn{1}{c|}{}                   & \multicolumn{5}{c|}{RCVQA}                                                                                                                                                                                                                                                                   & \multicolumn{6}{c}{RCID}                                                                                                                                                                                                          \\ \midrule
\multicolumn{1}{c|}{}                   & \multicolumn{1}{l}{}                      & \multicolumn{1}{l}{}                      &                                                                     &                                                                     & \multicolumn{1}{l|}{}                                   &                                      &                                      &                                       &                                       &                                  &                                  \\ 
\multicolumn{1}{c|}{\multirow{-3}{*}{}} 
& \multicolumn{1}{l}{\multirow{-2}{*}{V2C}} 
& \multicolumn{1}{l}{\multirow{-2}{*}{C2V}} 
& \multirow{-2}{*}{\begin{tabular}[c]{@{}c@{}}UA\\ V2C\end{tabular}} 
& \multirow{-2}{*}{\begin{tabular}[c]{@{}c@{}}UA\\ C2V\end{tabular}} 
& \multicolumn{1}{l|}{\multirow{-2}{*}{\textbf{Overall}}} 
& \multirow{-2}{*}{{CC}}
& \multirow{-2}{*}{{CA}}
& \multirow{-2}{*}{{CVC}}
& \multirow{-2}{*}{\textbf{RCIDS\textsubscript{ctx}}} 
& \multirow{-2}{*}{\textbf{CLIPS}} 
& \multirow{-2}{*}{\textbf{RCIDS}} \\ \midrule
\
\textbf{Proprietary}                    &                                           &                                           &                                                                     &                                                                     &                                                         &                                      &                                      &                                       &                                       &                                  &                                  \\
Gemini Flash 1.5 8B                     & 58.73                                     & 37.64                                     & 68.61                                                               & 25.48                                                               & 47.61                                                   & 20.88                                & 26.97                                & 26.77                                 & 24.87                                 & 87.11                            & 38.70                            \\
Gemini Flash 1.5                        & 70.43                                     & 42.76                                     & 70.88                                                               & 43.4                                                                & 56.87                                                   & 11.75                                & 17.24                                & 18.48                                 & 15.82                                 & 90.25                            & 26.93                            \\
Gemini Flash 2.0                        & 59.49                                     & 40.65                                     & 76.56                                                               & 41.15                                                               & 54.46                                                   & 17.82                                & 22.22                                & 20.94                                 & 20.33                                 & 84.76                            & 32.79                            \\
GPT-4o-mini                             & 49.60                                     & 28.95                                     & 56.34                                                               & 33.15                                                               & 42.01                                                   & 44.25                                & 63.18                                & 58.26                                 & 55.23                                 & 78.00                            & 64.67                            \\ \midrule
\textbf{Open-source}                    &                                            &                                           &                                                                     &                                                                     &                                                         &                                      &                                      &                                       &                                       &                                  &                                  \\
InternVL2-2B                            & 51.19                                     & 30.4                                      & 13.18                                                               & 1.01                                                                & 23.94                                                   & 18.72                                & 29.27                                & 22.86                                 & 23.62                                 & {\ul 83.70}                      & 36.84                            \\
InternVL2-4B                            & 49.26                                     & 32.52                                     & 36.44                                                               & 9.24                                                                & 31.87                                                   & 55.03                                & 59.49                                & 56.50                                 & 57.01                                 & 83.20                            & 67.66                            \\
InternVL2-8B                            & 52.85                                     & 35.30                                     & 39.57                                                               & 20.97                                                               & 37.17                                                   & 52.95                                & 61.41                                & 59.20                                 & 57.85                                 & \textbf{83.71}                   & 68.42                            \\
MiniCPM V2.6 (8B)                      & 51.90                                     & 33.18                                     & 15.58                                                               & 4.51                                                                & 26.30                                                   & 40.95                                & 48.43                                & 42.96                                 & 44.12                                 & 80.67                            & 57.04                            \\\midrule
Qwen2-VL 2B & 25.80 & 9.47 & 24.81 & 4.51 & 16.15 & 16.30 & 26.28 & 20.58 & 21.05 & 55.78 & 30.57 \\
\rowcolor[HTML]{E3F0FD}
RC-Qwen2-VL 2B (Ours) & \underline{70.95} \raisebox{-.5ex}{\scriptsize\textcolor[HTML]{228B22}{+45.15}} & \underline{50.45} \raisebox{-.5ex}{\scriptsize\textcolor[HTML]{228B22}{+40.98}} & \underline{75.51} \raisebox{-.5ex}{\scriptsize\textcolor[HTML]{228B22}{+50.70}} & \underline{95.83} \raisebox{-.5ex}{\scriptsize\textcolor[HTML]{228B22}{+91.32}} & \underline{73.18} \raisebox{-.5ex}{\scriptsize\textcolor[HTML]{228B22}{+57.03}} & \underline{97.38} \raisebox{-.5ex}{\scriptsize\textcolor[HTML]{228B22}{+81.08}} & \textbf{97.09} \raisebox{-.5ex}{\scriptsize\textcolor[HTML]{228B22}{+70.81}} & \underline{81.65} \raisebox{-.5ex}{\scriptsize\textcolor[HTML]{228B22}{+61.07}} & \underline{92.04} \raisebox{-.5ex}{\scriptsize\textcolor[HTML]{228B22}{+70.99}} & 68.67 \raisebox{-.5ex}{\scriptsize\textcolor[HTML]{228B22}{+12.89}} & \underline{78.66} \raisebox{-.5ex}{\scriptsize\textcolor[HTML]{228B22}{+48.09}} \\

Qwen2-VL 7B & 49.49 & 34.86 & 20.67 & 14.54 & 29.89 & 28.89 & 32.66 & 27.05 & 29.53 & 83.08 & 43.58 \\
\rowcolor[HTML]{E3F0FD}
RC-Qwen2-VL 7B (Ours) & \textbf{80.01} \raisebox{-.5ex}{\scriptsize\textcolor[HTML]{228B22}{+30.52}} & \textbf{59.24} \raisebox{-.5ex}{\scriptsize\textcolor[HTML]{228B22}{+24.38}} & \textbf{85.96} \raisebox{-.5ex}{\scriptsize\textcolor[HTML]{228B22}{+65.29}} & \textbf{97.86} \raisebox{-.5ex}{\scriptsize\textcolor[HTML]{228B22}{+83.32}} & \textbf{80.77} \raisebox{-.5ex}{\scriptsize\textcolor[HTML]{228B22}{+50.88}} & \textbf{97.98} \raisebox{-.5ex}{\scriptsize\textcolor[HTML]{228B22}{+69.09}} & \underline{95.17} \raisebox{-.5ex}{\scriptsize\textcolor[HTML]{228B22}{+62.51}} & \textbf{85.85} \raisebox{-.5ex}{\scriptsize\textcolor[HTML]{228B22}{+58.80}} & \textbf{93.00} \raisebox{-.5ex}{\scriptsize\textcolor[HTML]{228B22}{+63.47}} & 73.92 \raisebox{-.5ex}{\scriptsize\textcolor[HTML]{D2042D}{-9.16}} & \textbf{82.37} \raisebox{-.5ex}{\scriptsize\textcolor[HTML]{228B22}{+38.79}} \\
\bottomrule
\end{tabular}}
\caption{Experimental results (\%) on region-level context-aware multimodal tasks. V2C and C2V denote region-level visual-to-context VQA and region-level context-to-visual VQA, respectively. UA stands for unanswerable. CC, CA, and CVC represent the contextual coverage score, contextual accuracy score, and context-visual consistency score, respectively. Among open-source models, the best results are highlighted in \textbf{bold}, while the second-best results are {\ul underlined}. RC-Qwen2-VL models not only lead their similarly sized open-source counterparts but also outperform certain closed-source models.}
\label{tab1}

\end{table*}

\begin{table*}[]
\centering
\resizebox{\textwidth}{!}{ 
\begin{tabular}{l|c|ccc|ccc|ccc}
\toprule
Model             & Type                                  & \multicolumn{3}{c|}{Closed-World} & \multicolumn{3}{c|}{Open-World}  & \multicolumn{3}{c}{Oracle}       \\
                  &                                       & V2C    & C2V   & \textbf{Overall} & V2C   & C2V   & \textbf{Overall} & V2C   & C2V   & \textbf{Overall} \\\midrule
GPT-4o            & \multirow{2}{*}{\textbf{Proprietary}} & 48.79  & 44.42 & 46.75            & 64.96 & 65.44 & 65.18            & 71.75 & 70.76 & 71.29            \\
Gemini Pro 2.5    &                                       & 68.99  & 68.07 & 68.56            & 65.36 & 67.21 & 66.23            & 78.77 &79.76 & 79.23            \\\midrule
InternVL2-2B      & \multirow{6}{*}{\textbf{Open-source}} & 30.72  & 25.10 & 28.10            & 42.58 & 29.24 & 36.35            & 39.99 & 38.44 & 39.26            \\
InternVL2-4B      &                                       & 29.80  & 23.98 & 27.09            & 51.78 & 53.22 & 52.45            & 41.08 & 49.61 & 45.06            \\
InternVL2-8B      &                                       & 35.39  & 27.73 & \underline{31.81}            & \underline{48.73} & \underline{54.99} & \underline{51.66}            & \underline{60.47} & \underline{71.48} & \underline{65.61}            \\
MiniCPM V2.6 (8B) &                                       & 34.81  & 25.95 & 30.67            & 49.48 & 42.05 & 46.01            & 51.32 & 56.64 & 53.80            \\
RAP-LLaVA 13B     & & 5.12   & 4.80  & 4.97             & 6.16  & 6.11  & 6.13             & 7.25  & 7.29  & 7.27             \\\midrule
Qwen2-VL 2B & & 28.25 & 21.68 & 25.18 & 29.69 & 13.60 & 22.18 & 41.94 & 37.06 & 39.66 \\
\rowcolor[HTML]{E3F0FD}
RC-Qwen2-VL 2B (Ours) & & \underline{32.28} \raisebox{-.5ex}{\scriptsize\textcolor[HTML]{228B22}{+4.03}} & \underline{28.32} \raisebox{-.5ex}{\scriptsize\textcolor[HTML]{228B22}{+6.64}} & 30.43 \raisebox{-.5ex}{\scriptsize\textcolor[HTML]{228B22}{+5.25}} & 41.20 \raisebox{-.5ex}{\scriptsize\textcolor[HTML]{228B22}{+11.51}} & 50.66 \raisebox{-.5ex}{\scriptsize\textcolor[HTML]{228B22}{+37.06}} & 45.61 \raisebox{-.5ex}{\scriptsize\textcolor[HTML]{228B22}{+23.43}} & 49.71 \raisebox{-.5ex}{\scriptsize\textcolor[HTML]{228B22}{+7.77}} & 56.24 \raisebox{-.5ex}{\scriptsize\textcolor[HTML]{228B22}{+19.18}} & 52.76 \raisebox{-.5ex}{\scriptsize\textcolor[HTML]{228B22}{+13.10}} \\

Qwen2-VL 7B & & 34.06 & 28.06 & 31.26 & 44.65 & 34.17 & 39.75 & 54.95 & 59.13 & 56.90 \\
\rowcolor[HTML]{E3F0FD}
RC-Qwen2-VL 7B (Ours) & & \textbf{41.83} \raisebox{-.5ex}{\scriptsize\textcolor[HTML]{228B22}{+7.77}} &  \textbf{43.17} \raisebox{-.5ex}{\scriptsize\textcolor[HTML]{228B22}{+15.11}} &  \textbf{42.45} \raisebox{-.5ex}{\scriptsize\textcolor[HTML]{228B22}{+11.19}} &  \textbf{49.71} \raisebox{-.5ex}{\scriptsize\textcolor[HTML]{228B22}{+5.06}} &  \textbf{66.82} \raisebox{-.5ex}{\scriptsize\textcolor[HTML]{228B22}{+32.65}} &  \textbf{57.70} \raisebox{-.5ex}{\scriptsize\textcolor[HTML]{228B22}{+17.95}} &  \textbf{71.29} \raisebox{-.5ex}{\scriptsize\textcolor[HTML]{228B22}{+16.34}} &  \textbf{87.45} \raisebox{-.5ex}{\scriptsize\textcolor[HTML]{228B22}{+28.32}} &  \textbf{78.83} \raisebox{-.5ex}{\scriptsize\textcolor[HTML]{228B22}{+21.93}} \\\bottomrule
\end{tabular}}
\caption{Experimental results (\%) on RC\&P-Bench. V2C and C2V denote Visual-to-Context VQA and Context-to-visual VQA. The best result in each setting is bold and the second is
underlined in the open-source models. }
\label{MPU}
\end{table*}

\begin{table}[t]

\centering

\vspace{-3mm}
\resizebox{\columnwidth}{!}{\begin{tabular}{l|c|c|ccc}
\toprule
{Method}  &{Train}&{\#Image} &Visual & Text-only &Weighted\\
\midrule
MyVLM-LLaVA (13B)   &\ding{51} & 5 & 91.18   & - & - \\
Yo'LLaVA (13B)      &\ding{51} & 5 & 92.94   & \underline{88.25} &{90.60} \\
RAP-LLaVA (13B)     &\ding{55} & 1 & {93.53} & \textbf{93.75} & \textbf{93.64} \\
RAP-Phi3-V (3.8B)   &\ding{55} & 1 & {94.12} & 85.00 & 89.56\\

\midrule
Qwen2-VL 2B  & \ding{55} & 0 & 95.29 & 85.00 & 90.15 \\
\rowcolor[HTML]{E3F0FD}
RC-Qwen2-VL 2B (Ours) & \ding{55} & 0 & 95.29 & 85.25 \raisebox{-.5ex}{\scriptsize\textcolor[HTML]{228B22}{(+0.25)}} & 90.27 \raisebox{-.5ex}{\scriptsize\textcolor[HTML]{228B22}{(+0.12)}} \\

Qwen2-VL 7B & \ding{55} & 0 & \underline{96.47} & 86.25 & 91.36 \\
\rowcolor[HTML]{E3F0FD}
RC-Qwen2-VL 7B (Ours) & \ding{55} & 0 & \textbf{97.06} \raisebox{-.5ex}{\scriptsize\textcolor[HTML]{228B22}{(+0.59)}} & 86.50 \raisebox{-.5ex}{\scriptsize\textcolor[HTML]{228B22}{(+0.25)}} & \underline{91.78} \raisebox{-.5ex}{\scriptsize\textcolor[HTML]{228B22}{(+0.42)}} \\

\bottomrule
\end{tabular}}
\caption{Experimental results (\%) on personalized visual question answering. The best result in each setting is bold and the second is underlined. Weighted results are computed as arithmetic means.}
\label{PQA}
\end{table}

\begin{table}[]
\centering
\resizebox{\columnwidth}{!}{

\begin{tabular}{l|ccc|ccc}
\toprule
\multicolumn{1}{c|}{}                            & \multicolumn{3}{c|}{RCVQA}                               & \multicolumn{3}{c}{RCID}                                 \\ \midrule
\multicolumn{1}{c|}{\multirow{-2}{*}{\textbf{}}} & {R} & {P} & \textbf{F1} & {R} & {P} & \textbf{F1} \\ \midrule
\textbf{Proprietary}                             &                 &                    &                   &                 &                    &                   \\
Gemini Flash 1.5 8B                              & 77.13           & 69.36              & 73.04             & 77.82           & 10.50              & 18.50              \\
Gemini Flash 1.5                                 & 87.40            & 83.62              & 85.47             & 90.74           & 1.31               & 2.58              \\
Gemini Flash 2.0                                 & 72.69           & 57.89              & 64.45             & 85.73           & 14.47              & 24.76             \\
GPT-4o-mini                                      & 63.67           & 66.47              & 65.04             & 55.07           & 72.15              & 62.46             \\ \midrule
\textbf{Open-source}                             &                 &                    &                   &                 &                    &                   \\
InternVL2-2B                                     & 1.24            & 55.93              & 2.44              & 4.17            & \textbf{62.73}     & 7.82              \\
InternVL2-4B                                     & 26.25           & 72.25              & 38.51             & 14.17           & {\ul 59.94}        & 22.92             \\
InternVL2-8B                                     & 60.13           & 80.23              & 68.74             & 24.92           & 47.95              & 32.80              \\
MiniCPM V2.6 (8B)                               & 74.20            & 74.67              & 74.43             & 51.32           & 49.44              & {\ul 50.37}       \\\midrule
Qwen2-VL-2B & 15.31 & 0.22 & 0.43 & 44.78 & 18.55 & 26.24 \\
\rowcolor[HTML]{E3F0FD}
RC-Qwen2-VL 2B (Ours) & \underline{89.51} \raisebox{-.5ex}{\scriptsize\textcolor[HTML]{228B22}{+74.20}} & \underline{93.46} \raisebox{-.5ex}{\scriptsize\textcolor[HTML]{228B22}{+93.24}} & \underline{91.45} \raisebox{-.5ex}{\scriptsize\textcolor[HTML]{228B22}{+91.02}} & 44.25 \raisebox{-.5ex}{\scriptsize\textcolor[HTML]{D2042D}{-0.53}} & 48.41 \raisebox{-.5ex}{\scriptsize\textcolor[HTML]{228B22}{+29.86}} & 46.23 \raisebox{-.5ex}{\scriptsize\textcolor[HTML]{228B22}{+19.99}} \\

Qwen2-VL-7B & 61.15 & 66.58 & 63.75 & 30.81 & 21.81 & 25.54 \\
\rowcolor[HTML]{E3F0FD}
RC-Qwen2-VL 7B (Ours) & \textbf{92.57} \raisebox{-.5ex}{\scriptsize\textcolor[HTML]{228B22}{+31.42}} & \textbf{96.21} \raisebox{-.5ex}{\scriptsize\textcolor[HTML]{228B22}{+29.63}} & \textbf{94.35} \raisebox{-.5ex}{\scriptsize\textcolor[HTML]{228B22}{+30.60}} & \textbf{61.05} \raisebox{-.5ex}{\scriptsize\textcolor[HTML]{228B22}{+30.24}} & 53.82 \raisebox{-.5ex}{\scriptsize\textcolor[HTML]{228B22}{+32.01}} & \textbf{57.21} \raisebox{-.5ex}{\scriptsize\textcolor[HTML]{228B22}{+31.67}} \\
\bottomrule
\end{tabular}}
\caption{Experimental results (\%) of citation generation on the RCVQA and RCID tasks. Among open-source models, the best results are highlighted in \textbf{bold}, while the second-best results are {\ul underlined}. RC-Qwen2-VL models lead their similarly sized open-source counterparts and outperform certain closed-source models.}
\label{tab2}
\end{table}

\section{Experiments}

\subsection{Training Details} We fine-tuned the Qwen2-VL-2B-Instruct and Qwen2-VL-7B-Instruct models using the LoRA method (rank = 8, alpha = 16), while keeping the visual encoder frozen. Training was conducted for 20,000 steps with a batch size of 32 and a learning rate of 5.0e-5, utilizing a cosine learning rate scheduler with a warmup ratio of 0.05. The training process was performed on 2 NVIDIA RTX 3090 GPUs in bfloat16 precision to optimize computational efficiency.

\subsection{Experimental Results on RCMU Tasks}
\label{e2}

\textbf{Settings.} We evaluate the models‘ region-level context-aware multimodal (RCMU) capabilities on the RCMU test set, comprising three tasks: Region-based Contextual Image Description (RCID), Region-based Contextual Visual Question Answering (RCVQA), and Multimodal Context-aware Content Generation (MCCG). We compare our models against state-of-the-art MLLMs, including proprietary models like Gemini Flash 1.5 8B \cite{Reid2024Gemini1U}, Gemini Flash 1.5 \cite{Reid2024Gemini1U}, Gemini Flash 2.0 \cite{deepmind2024gemini}, GPT-4o-mini \cite{hurst2024gpt} and open-source models such as InternVL2 series  \cite{chen2024far, Chen2023InternVS}, Qwen2-VL series  \cite{Wang2024Qwen2VLEV}) and MiniCPM V2.6 \cite{yao2024minicpm}. For the RCID task, we employed RCIDScore to provide a comprehensive and fine-grained evaluation. The results from traditional metrics, such as BLEU \cite{papineni2002bleu}, ROUGE-L \cite{lin2004rouge}, and CIDEr \cite{vedantam2015cider}, are included in the appendix for reference. For the RCVQA task, we used the DeepSeek V3 \cite{liu2024deepseek} model to assess the correctness of the answers. For the MCCG task, following \cite{gao2023enabling}, we evaluated the recall, precision, and F1 score of citations. Specifically, citation recall refers to the percentage of sentences/responses that can be supported by the corresponding cited passages. Citation precision indicates the percentage of citations in the response that effectively support their respective sentences. The citation F1 score is the harmonic mean of citation recall and citation precision. We compute citation recall and citation precision following the approach outlined in \cite{gao2023enabling}, with one key difference: we consider images as a default source of information that does not require explicit citation. As a result, each sentence inherently has an implicit citation to the image, and we take this implicit citation into account during evaluation. We used Gemini Flash 1.5 \cite{Reid2024Gemini1U} to determine whether a response is supported by the corresponding cited passages and images. The prompts used in the evaluation are displayed in the appendix.
\paragraph{Results on RCVQA and RCID}
The results on RCVQA and RCID tasks are shown in Table \ref{tab1}. For the RCVQA task, the RC-Qwen2-VL 2B and RC-Qwen2-VL 7B models significantly outperform all compared open-source and closed-source models, achieving an accuracy improvement of over 50\% compared to the baseline Qwen-VL 2B and Qwen-VL 7B models. Notably, while most open-source models struggle with unanswered questions, RC-Qwen2-VL models demonstrate superior capability in addressing these challenges. For the RCID task, the RC-Qwen2-VL models achieved the highest RCIDScore, surpassing all the compared open-source and closed-source models, demonstrating exceptional capabilities in region-level context-aware image description generation. The descriptions generated by the RC-Qwen2-VL models showed significant improvements over the baseline models in terms of context coverage, accuracy, and consistency between textual and visual information. Furthermore, we observed that after RCVIT, the visual quality of descriptions generated by the RC-Qwen2-VL 7B model slightly declined. This may be attributed to the difficulty of balancing a comprehensive description of visual content with contextual relevance and fluency. In summary, the RC-Qwen2-VL models outperform both open-source and closed-source models in RCVQA and RCID, demonstrating strong RCMU capability.

\paragraph{Results on Multimodal Contextual Citation Generation}
We assessed the quality of the citations generated by models in both the RCVQA and RCID tasks. Given the distinct nature of these tasks, RCVQA requires a citation to be appended at the end of the answer, while RCID involves providing citations at multiple relevant points throughout the description. The results are shown in Table \ref{tab2}. In the RCVQA task, the RC-Qwen2-VL models demonstrate the highest citation quality among all open-source and closed-source models in the comparison, achieving top performance in citation recall, citation precision, and F1 score. When compared to the baseline models, the RC-Qwen2-VL models show significant improvements. For example, RC-Qwen2-VL 2B achieves an F1 score improvement of 91.02\% compared to the baseline models, while RC-Qwen2-VL 7B demonstrates an improvement of 30.6\%. In the RCID task, the RC-Qwen2-VL models achieve the highest citation recall and F1 score among open-source models, with the F1 score of RC-Qwen2-VL 7B matching the top performance of closed-source models. Compared to the baseline models, the RC-Qwen2-VL models demonstrate significant improvements, with the F1 score of RC-Qwen2-VL 2b rising by 19.99\% and the F1 score of RC-Qwen2-VL 7b increasing by 31.67\%. Overall, the RC-Qwen2-VL models excel in multimodal context citation generation tasks.

\begin{table}[]
\centering
\resizebox{\columnwidth}{!}{
\begin{tabular}{llcccc}
\toprule
\textbf{}                                                                                         & \textbf{}                                    & \multicolumn{2}{c}{\begin{tabular}[c]{@{}c@{}}RC-Qwen2-VL\\ 2B\end{tabular}} & \multicolumn{2}{c}{\begin{tabular}[c]{@{}c@{}}RC-Qwen2-VL\\ 7B\end{tabular}} \\ \midrule
\multicolumn{2}{c|}{Citation Annotation}                                                                                                               & wo/                        & \multicolumn{1}{c|}{w/}                         & wo/                                   & w/                                   \\ \midrule
\multicolumn{1}{l|}{\multirow{5}{*}{RCVQA}}                                                       & \multicolumn{1}{l|}{V2C Acc}                 & 70.77                      & \multicolumn{1}{c|}{70.95}                      & 81.33                                 & 80.01                                \\
\multicolumn{1}{l|}{}                                                                             & \multicolumn{1}{l|}{C2V Acc}                 & 50.56                      & \multicolumn{1}{c|}{50.45}                      & 60.80                                 & 59.24                                \\
\multicolumn{1}{l|}{}                                                                             & \multicolumn{1}{l|}{UA V2C Acc}             & 58.97                      & \multicolumn{1}{c|}{75.51}                      & 84.01                                 & 85.96                                \\
\multicolumn{1}{l|}{}                                                                             & \multicolumn{1}{l|}{UA C2V Acc}             & 94.81                      & \multicolumn{1}{c|}{95.83}                      & 95.94                                 & 97.86                                \\
\multicolumn{1}{l|}{}                                                                             & \multicolumn{1}{l|}{\textbf{Overall Acc}}    & 68.78                      & \multicolumn{1}{c|}{\textbf{73.18}}             & 80.52                                 & \textbf{80.77}                       \\ \midrule
\multicolumn{1}{l|}{\multirow{3}{*}{\begin{tabular}[c]{@{}l@{}}Citation\\ in RCVQA\end{tabular}}} & \multicolumn{1}{l|}{Recall}                  & 0                          & \multicolumn{1}{c|}{89.51}                      & 6.07                                  & 92.57                                \\
\multicolumn{1}{l|}{}                                                                             & \multicolumn{1}{l|}{Precision}               & NA                         & \multicolumn{1}{c|}{93.46}                      & 85.87                                 & 96.21                                \\
\multicolumn{1}{l|}{}                                                                             & \multicolumn{1}{l|}{\textbf{F1 Score}}       & NA                         & \multicolumn{1}{c|}{\textbf{91.45}}             & 11.34                                 & \textbf{94.35}                       \\ \midrule
\multicolumn{1}{l|}{\multirow{6}{*}{RCID}}                                                        & \multicolumn{1}{l|}{CC}                      & 94.44                      & \multicolumn{1}{c|}{97.38}                      & 95.67                                 & 97.98                                \\
\multicolumn{1}{l|}{}                                                                             & \multicolumn{1}{l|}{CA}                      & 93.95                      & \multicolumn{1}{c|}{97.09}                      & 95.46                                 & 95.17                                \\
\multicolumn{1}{l|}{}                                                                             & \multicolumn{1}{l|}{CVC}                     & 83.65                      & \multicolumn{1}{c|}{81.65}                      & 86.13                                 & 85.85                                \\
\multicolumn{1}{l|}{}                                                                             & \multicolumn{1}{l|}{\textbf{Context Score}}  & 90.68                      & \multicolumn{1}{c|}{\textbf{92.04}}             & 92.42                                 & \textbf{93.00}                       \\
\multicolumn{1}{l|}{}                                                                             & \multicolumn{1}{l|}{\textbf{LLM2Clip Score}} & \textbf{70.05}             & \multicolumn{1}{c|}{68.67}                      & \textbf{77.28}                        & 73.92                                \\
\multicolumn{1}{l|}{}                                                                             & \multicolumn{1}{l|}{\textbf{Final Score}}    & \textbf{79.04}             & \multicolumn{1}{c|}{78.66}                      & \textbf{84.18}                        & 82.37                                \\ \midrule
\multicolumn{1}{l|}{\multirow{3}{*}{\begin{tabular}[c]{@{}l@{}}Citation\\ in RCID\end{tabular}}}  & \multicolumn{1}{l|}{Recall}                  & 9.90                       & \multicolumn{1}{c|}{44.25}                      & 14.84                                 & 61.05                                \\
\multicolumn{1}{l|}{}                                                                             & \multicolumn{1}{l|}{Precision}               & 47.13                      & \multicolumn{1}{c|}{48.41}                      & 18.50                                 & 53.82                                \\
\multicolumn{1}{l|}{}                                                                             & \multicolumn{1}{l|}{\textbf{F1 Score}}       & 16.37                      & \multicolumn{1}{c|}{\textbf{46.23}}             & 16.47                                 & \textbf{57.21}                      \\\bottomrule
\end{tabular}}
\caption{Ablation study on the impact of citation annotations in training data. Best results are highlighted in \textbf{bold}. UA stands for unanswerable. }
\label{tab3}
\end{table}

\begin{table}[]
\centering
\resizebox{\columnwidth}{!}{
\begin{tabular}{c|cccc
>{\columncolor[HTML]{E3F0FD}}c }
\toprule
P/N Ratio & V2C   & C2V   & UA V2C & UA C2V & \textbf{Overall} \\\midrule
wo/ UA data   & 69.63 & 50.45 & 1.95    & 0.11    & 30.54            \\
1:1       & 67.45 & 47.33 & 68.74   & 94.59   & \textbf{69.53}   \\
2:1       & 68.43 & 46.99 & 63.56   & 92.11   & 67.77            \\
4:1       & 69.07 & 48.22 & 54.34   & 87.03   & 64.67            \\
1:2       & 65.98 & 43.10 & 70.42   & 95.72   & 68.80           \\\bottomrule
\end{tabular}}
\caption{Ablation study on the impact of no-answer RCVQA data proportion in training on RCVQA performance. Models here were trained for 5000 steps. UA stands for unanswerable.  }
\label{tab4}
\end{table}

\begin{table}[]
\centering
\begin{tabular}{l|c|c}
\toprule
        & Ref & $\tau$    \\\midrule
BLEU-1  & \ding{51}      & 25.55  \\
BLEU-4  & \ding{51}      & 28.54  \\
ROUGE-L & \ding{51}      & 29.71 \\
METEOR  & \ding{51}      & 47.42  \\
CIDEr   & \ding{51}     & 50.68  \\\midrule
CLIPS   &\ding{55}      & 0.19 \\
RCIDS(ours)   & \ding{55}     & \textbf{51.84}\\\bottomrule 
\end{tabular}
\caption{Correlations with human
judgment.}
\label{correlation}
\end{table}

\begin{figure}[!t]
\begin{center}
\centerline{\includegraphics[width=\columnwidth]{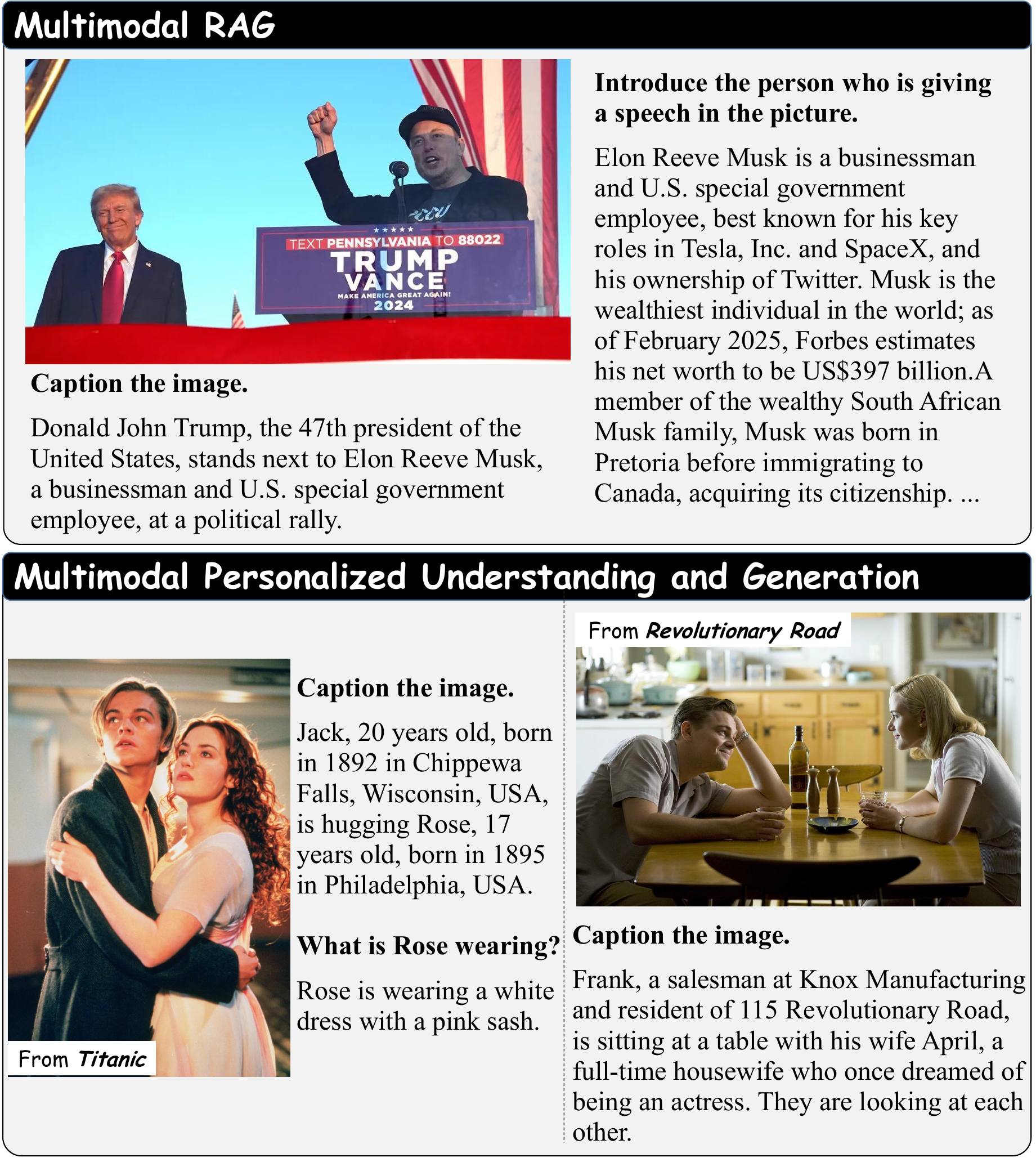}}
\caption{Showcase of our models in practical applications. Amazon Rekognition Service is used here for celebrity/personalized detection and recognition.}
\label{fig5}
\end{center}
\vspace{-20pt}

\end{figure}

\subsection{Experimental Results on Personalized Multimodal Understanding}
\label{e4}
\textbf{Settings}. We evaluate the model's personalized multimodal understanding capabilities on RC\&P-Bench and the personalized question answering benchmark introduced by Yo’LLaVA \cite{nguyen2024yo}. We follow the setup in the RAP \cite{hao2025rap} and employ a multimodal retriever. For these tasks, we use accuracy as the evaluation metric. We compare our models against state-of-the-art MLLMs, including proprietary models like Gemini Pro 2.5 \cite{comanici2025gemini}, GPT-4o \cite{hurst2024gpt} and open-source models such as InternVL2 series \cite{chen2024far, Chen2023InternVS}, Qwen2-VL series \cite{Wang2024Qwen2VLEV}, MiniCPM V2.6 \cite{yao2024minicpm} and specialized personalized multimodal understanding models like MyVLM \cite{alaluf2024myvlm}, Yo'LLaVA \cite{nguyen2024yo} and RAP \cite{hao2025rap}. We evaluated accuracy across three distinct settings. In the \textbf{Closed-World} setting, we assess the overall system's accuracy, regardless of whether the correct information was retrieved. The \textbf{Open-World} setting, in contrast, considers 'insufficient information' a correct response when the necessary information is not retrieved. Finally, the \textbf{Oracle} setting bypasses retrieval by directly providing gold evidence, thereby evaluating only the answer generation capability.

\textbf{Results and Analysis}. 
The results on RC\&P-Bench are presented in Table \ref{MPU}, showing that the RC-Qwen2-VL models achieve significant improvements over the baseline models. Specifically, RC-Qwen2-VL 2B shows a remarkable 23.43 boost in the Open-World Overall score and 13.10 in the Oracle setting compared to its Qwen2-VL 2B baseline. The gains are even more pronounced for the larger model. RC-Qwen2-VL 7B surpasses its baseline by 17.95 in the Open-World Overall score and an impressive 21.93 in the Oracle setting. The most substantial individual improvements are seen in the Context-to-Visual (C2V) tasks, with RC-Qwen2-VL 7B gaining 32.65 in the Open-World C2V and 28.32 in the Oracle C2V tasks. Furthermore, the RC-Qwen2-VL 7B model has achieved state-of-the-art (SOTA) performance among open-source models and is comparable to or even surpasses top-tier proprietary models. For instance, in the Oracle setting, RC-Qwen2-VL 7B's Overall score of 78.83\% significantly outperforms other open-source models like InternVL2-8B (65.61\%) and MiniCPM V2.6 (53.80\%). Notably, this score also surpasses the proprietary GPT-4o (71.29\%) and is highly competitive with Gemini Pro 2.5 (79.23\%). This indicates that the capabilities of the RC-Qwen2-VL models, developed through region-level context-aware visual instruction tuning, can be effectively transferred to multimodal personalized understanding.

The results on the personalized question answering benchmark are presented in Table \ref{PQA}. Our proposed models demonstrate exceptional performance and efficiency. Specifically, the RC-Qwen2-VL 7B model sets a new state-of-the-art record on the visual personalized question answering task with a score of 97.06\%. Critically, this is achieved in a zero-shot setting without requiring any extra personalized training, a stark contrast to other methods that depend on personalized tuning. Compared to its baseline, our method delivers consistent improvements across all metrics. The 7B model achieves a weighted average score of 91.78\%, securing the second-highest rank among all methods and proving its strong competitiveness despite its smaller model size and more challenging setup.

\subsection{Further Analyses}
\label{e3}
\paragraph{Ablation Study}
We explored the effect of citation annotations in the training data on the region-level context-aware ability of the model. Specifically, we compared the model's performance on training data with and without citation annotations. The experimental results are presented in Table \ref{tab3}. As shown, incorporating citation annotations into the training data significantly improves the quality of the generated citations and modestly enhances the model's region-level context-aware multimodal understanding capability. However, this comes at the cost of reduced visual quality in the generated image descriptions.

We also investigated how varying the proportion of no-answer RCVQA data in the training set impacts the model’s RCVQA performance. Specifically, we trained five different RC-Qwen2-VL variants for 5000 steps each, under various conditions: one with no no-answer RCVQA data, and others with different ratios of answerable to unanswerable RCVQA data (1:1, 2:1, 4:1, 1:2). The results are presented in Table \ref{tab4}. As shown, incorporating unanswerable RCVQA data into the training set significantly enhances the model's performance on such questions. Furthermore, the model achieves its highest overall accuracy on the RCVQA task when the ratio of answerable to unanswerable RCVQA training data is 1:1.

To evaluate the consistency of image captioning metrics with human subjective judgment on Region-level context-aware image descriptions, we manually scored 500 (image, description) pairs. The scoring was based on a five-level scale: \textbf{Excellent:} No important visual or contextual information is missing; \textbf{Good:} Minor omissions of visual or contextual information; \textbf{Average:} Important visual or contextual information is missing; \textbf{Poor:} Either the visual or the contextual information is completely omitted; \textbf{Very Poor:} The caption is not acceptable due to grammatical errors, incomplete sentences, or other flaws.
The experiment measures this consistency by calculating the Kendall's Tau ($\tau$) correlation coefficient between each metric's scores and the human ratings. A higher correlation coefficient indicates that the results of the automatic evaluation metric align more closely with the standards of human judgment. The results are shown in Table \ref{correlation}, The proposed reference-free metric, RCIDS, achieves the highest correlation score (51.84), demonstrating a stronger alignment with human assessment than all other tested metrics. Notably, it not only vastly outperforms the other reference-free metric, CLIPS (0.19), but also surpasses established reference-based standards like CIDEr (50.68), proving its state-of-the-art performance and utility in evaluating text quality without needing a ground truth reference.


\section{Conclusion}

This paper introduces Region-level Context-aware Multimodal Understanding (RCMU), a new task requiring models to respond to instructions by integrating image content with textual information about specific regions or objects. The paper proposes Region-level Context-aware Visual Instruction Tuning (RCVIT) to equip MLLMs with this capability, along with the RCMU dataset for training and evaluation, and RCIDScore, a novel reference-free evaluation metric. Experiments show that models trained with RCVIT and the RCMU dataset outperform existing MLLMs on RCMU tasks and demonstrate applications in multimodal RAG and personalized conversation.

\bibliographystyle{IEEEtran}

\bibliography{main}

\end{document}